\documentclass[11pt]{article}
\usepackage[letterpaper,margin=1in]{geometry}
\usepackage[utf8]{inputenc}
\usepackage{amsmath,amssymb}
\usepackage{booktabs}
\usepackage{graphicx}
\usepackage[authoryear,square]{natbib}
\usepackage[hidelinks]{hyperref}
\usepackage{url}

\hypersetup{
  pdftitle={Momentum, Commentary Tone, and Earnings Surprise in LLM Financial Forecasts: A Controlled Study of Six Q8 Checkpoint Conversions},
  pdfauthor={Alina Khaybullina}
}

\title{Momentum, Commentary Tone, and Earnings Surprise\\in LLM Financial Forecasts\\[4pt]
\large A Controlled Study of Six Q8 Checkpoint Conversions}
\author{Alina Khaybullina}
\date{October 2026}

\begin{document}
\maketitle

\begin{abstract}
\noindent Which inputs drive a language model's financial forecast? We construct synthetic earnings announcements that independently reverse 20-day momentum, non-informational commentary tone, and whether fixed actual EPS beats a manipulated consensus. Six instruction/chat checkpoint conversions from four families (0.49--7.61B base parameters) receive 2,016 prompts each under a direct-answer Q8\_0 protocol. The design includes 24 numerical blocks, six commentary pairs, profits and losses, rising and falling annual EPS, and two surprise sizes. Both A/B answer assignments and U/D cover all 384 primary scenarios per conversion. We estimate matched changes in conditional answer-prefix scores, not calibrated market probabilities. Across these checkpoints, the primary EPS contrast ranges from $-0.19$ to 86.80 percentage points. The magnitude of the primary signed tone contrast exceeds that of EPS in 5 of six models and momentum exceeds it in 5. Natural-word numerical diagnostic accuracy ranges from 51.0 to 95.8 percent. Answer-encoding, wording, context, and conversion controls qualify these results. The experiment measures observable input reliance and numerical recognition; it establishes neither an internal reasoning mechanism nor realized return predictability. Public benchmark prompts, recorded scores, and an interactive recorded-results explorer accompany the study.
\end{abstract}

\section{Introduction}
Which information actually changes a language model's market prediction? A plausible financial forecast may follow recent returns, react to the emotional framing of an announcement, or compare newly released earnings with expectations. Natural financial text usually combines these signals. If they all agree, the final forecast cannot distinguish their influence.

An earnings announcement makes the distinction concrete. Higher earnings than a year earlier can still miss consensus; a loss can beat an even worse expected loss. Upbeat commentary can accompany either numerical outcome. We hold actual earnings fixed and reverse the surprise by changing the stated consensus, while independently reversing momentum and commentary. This identifies the effect of the information supplied to the model under the tested protocol.

This paper expands an audited two-checkpoint pilot. The primary experiment now covers six checkpoints in four families, a wider size range, six commentary pairs, losses, annual declines, and small as well as large surprises. Label reversal is part of every primary case. The pilot's raw results are retained separately, and a conversion/runtime bridge compares saved full-precision scores with the new inference setup. The original expanded protocol was internally frozen before expanded inference, after the pilot had already been inspected; it is not a public preregistration or a fresh confirmation of pilot-independent hypotheses. A later coverage audit added profitable cases to the smaller word and office controls. That explicitly exploratory amendment changes neither the frozen primary prompts nor the primary estimands.

The study has three questions. First, how much do momentum, tone, and numerical surprise change the requested forecast score? Second, do those effects survive changes in answer encoding and wording? Third, can the model recognize the beat/miss relation when asked directly? These questions concern observable behavior. A numerical effect is not a proof of reasoning, and a momentum effect is not automatically a mistake: price history may contain genuine information in real markets. We do not observe market outcomes in this experiment.

\section{Related work}
\citet{lopezlira} evaluate language-model headline assessments against market returns. That motivates scrutiny of input processing, while their predictive estimand differs from our controlled output-intervention estimand. \citet{turpin} show that generated explanations can omit influential prompt features, supporting the use of output interventions rather than self-reported explanations. \citet{gsmsymbolic} use systematic numerical variants to test mathematical reasoning; our financial construction similarly separates factual comparisons from framing. \citet{zhao} document prompt and answer-preference effects, motivating complete label reversal here. \citet{fincall} introduce a broader multimodal earnings-surprise benchmark; our balanced synthetic cases complement rather than replace real financial data. \citet{camef} use event counterfactuals in financial forecasting, while our interventions probe a fixed model. \citet{vamvourellis} examine financial sentiment classification under reasoning prompts, a different task from return prediction. \citet{finstar} distinguish deterministic financial assessment from stochastic prediction, which is central to separating our diagnostic from the forecast score.

The Qwen2.5 family is described by \citet{qwen25}; official checkpoint documentation describes the SmolLM2, TinyLlama, and Phi models \citep{smollm2,tinyllama,phi}. We make no claim that counterfactual financial evaluation is generally new. The contribution is the expectation-reversal design combined with full answer-mapping controls and a reproducible multi-family comparison of factual recognition and forecast sensitivity.

\section{Benchmark and protocol}
\subsection{Numerical blocks}
We generate 24 earnings blocks with NumPy's default random generator and seed 20261008. Blocks cross two signs of actual EPS, two signs of annual EPS change, and two surprise-gap fractions (2\% or 15\% of absolute actual EPS), with three numerical replicates per combination. Absolute actual EPS is sampled uniformly between \$0.80 and \$4.20 and rounded to cents. For profitability sign $p_b$, annual-change sign $h_b$, and a draw $u_b \sim U(0.08,0.25)$, prior-year EPS is $a_b - h_b|a_b|u_b$, rounded to cents. We describe annual change in EPS levels; percentage growth across negative earnings is not used.

For surprise sign $e\in\{-1,+1\}$, consensus is
\begin{equation}
c_{b,e} = a_b - e\,g_b, \qquad g_b=\max\bigl(0.01,\operatorname{round}(|a_b|f_b,2)\bigr).
\end{equation}
Actual and prior EPS remain fixed within a block. A beat means $a_b>c_{b,e}$ even when both are negative. Changing consensus manipulates the information supplied to the model, not actual investor expectations. Revenue meets consensus, guidance is unchanged, and no other news is released.

Momentum sign $m$ selects positive or negative 20-day pre-release returns of the same block-specific magnitude, sampled uniformly from 3--12\% and rounded to one decimal. The market is flat. Sentiment sign $s$ selects one of six positive/negative commentary pairs, explicitly described as opinion with no additional facts. Pair assignments are randomized and balanced across blocks (four blocks per pair). Three introductions are similarly assigned (eight blocks per introduction). Every block uses both section orders: facts--prices--commentary and commentary--prices--facts. Pair and introduction are between-block assignments; order and the three financial inputs are crossed within blocks.

The eight $(m,s,e)$ combinations and two orders give 16 semantic scenarios per block and 384 per model. All receive three answer encodings: A=positive/B=negative (AB0), the reversed assignment (AB1), and U=positive/D=negative (UD). Thus there are 1,152 primary prompts per model. Counts refer to repeated counterfactual model evaluations, not independent market events.

\subsection{Controls and expanded coverage}
The isolated numerical diagnostic crosses 24 blocks, two surprise signs, and two sentence forms: 96 scenarios. Each receives both A/B assignments and natural beat/miss prefixes, giving 288 diagnostic prompts. Commentary removal covers every block, both momentum and surprise signs, facts-first order, and both assignments (192 prompts, 96 averaged cases). The original natural-word and office controls selected the first complete block assigned to each commentary pair: 96 prompts each. A coverage audit after three completed expanded runs found that all six selected blocks were losses. An internally timestamped amendment adds the first profitable block for each pair, selected only from frozen block metadata. The resulting controls are balanced for profitability and pair: 192 word prompts and 192 office prompts (96 averaged office cases). Annual-change and gap-size balance is not guaranteed in this subset. The original primary, diagnostic, and neutral designs remain unchanged; these additional controls are exploratory.

The original freeze covers 1,824 cases per model; the amendment adds 192. Combined coverage is 2,016 prompts per model, or 12,096 across six models. Another 48 pilot bridge cases for each repeated Qwen checkpoint give 12,192 newly scored records. The 1,248 saved pilot records are distinct and are not pooled into primary estimates.

The six commentary pairs are authored variants from one construction process, not independent human authors or sentiment annotations. Numerical signs and gap sizes broaden scenario coverage, but the experiment still uses short synthetic disclosures rather than observed financial news. Table~\ref{tab:pairs} lists every pair; full prompts are supplied as JSONL.

\subsection{Models and inference}
All checkpoints use revision-pinned Q8\_0 GGUF conversions in llama-cpp-python 0.3.16, with the official Hugging Face tokenizer and chat template from pinned revisions. The repeated Qwen tokenizers retain their pilot revisions. Qwen conversions are first-party; SmolLM2 and Phi conversions are by bartowski, and TinyLlama's is by TheBloke. These provenance differences and quantization are disclosed rather than assumed immaterial. Pinned tokenizer revisions do not certify the source weight revision or source precision used by any downloaded conversion, including first-party files. The distributed manifests identify the actual conversion files and their SHA-256 hashes.

\begin{table}[t]
\centering
\begin{tabular}{lrlll}
\toprule
Checkpoint & Params (B) & Family & Tokenizer rev. & GGUF rev. \\
\midrule
Qwen2.5-0.5B & 0.49 & Qwen & 7ae55760 & 9217f5db \\
Qwen2.5-1.5B & 1.54 & Qwen & 989aa798 & 91cad511 \\
SmolLM2-1.7B & 1.71 & SmolLM2 & 31b70e2e & 1f034647 \\
TinyLlama-1.1B & 1.10 & TinyLlama & fe8a4ea1 & 52e7645b \\
Phi-3.5-mini & 3.82 & Phi & 2fe19245 & 6d70da17 \\
Qwen2.5-7B & 7.61 & Qwen & a09a3545 & bb5d59e0 \\
\bottomrule
\end{tabular}
\caption{Six instruction/chat checkpoints, all evaluated as Q8\_0 conversions. Revision columns are abbreviated; complete hashes and file digests are distributed. Parameter counts describe the base checkpoints and are rounded.}
\label{tab:models}
\end{table}

Inference uses CPU execution, four single-token threads, four or eight batch-prefill threads, a 512-token context, and no sampling. The initial inference batch size was 512; after an execution-environment replacement the remaining Qwen7 cases and independent audits use batch size 128 to reduce memory. Twelve uncached comparisons against saved Qwen7 cases verify the restored runtime and batch-size change before resumption; this is an execution change, not a change to the prompt design. Early Qwen evaluations used four batch threads; later evaluations use eight after a 16-case Qwen7 timing check found identical scores at both thread counts. Evaluation order is optimized by grouping shared official-token prefixes; frozen prompts and estimands are unchanged. Every complete input and requested continuation fits within the context. All rendered inputs and candidate boundaries are checked. Native GGUF segmentation agrees with official Hugging Face tokenization for four conversions. TinyLlama's native piece segmentation and Phi's native special-token whitespace handling differ from the official tokenizers. Every official vocabulary ID is verified against the conversion (32,000 for TinyLlama and 32,011 for Phi), and official Hugging Face token IDs are supplied directly, bypassing native segmentation. Phi's remaining padded vocabulary slots are not prompt/candidate tokens. Prompt and token-array hashes, raw candidate log probabilities, margins, requested-prefix mass, and top-token information are recorded. Cached and full-context evaluation are compared before each run. Twelve spread-out cases per model are also re-evaluated independently without that cache.

\subsection{Measured scores and primary estimand}
For prompt $x$ and positive/negative candidate token sequences $y_+,y_-$, let
\begin{equation}
\ell_j(x)=\sum_{t=1}^{|y_j|}\log P(y_{j,t}\mid x,y_{j,<t}),\qquad
q(x)=\frac{\exp\ell_+(x)}{\exp\ell_+(x)+\exp\ell_-(x)}.
\end{equation}
Continuations are tokenized as suffixes of the complete rendered prompt plus candidate, preserving the actual token boundary. Multi-token words receive full prefix likelihoods. No end-of-answer token is included; these are prefix comparisons, not probabilities of complete generated responses. Requested-prefix mass is $\exp\ell_++\exp\ell_-$, and is reported to reveal conditioning on unlikely answer prefixes. Natural-word candidates can have different token lengths; full-prefix likelihood retains output-word and length preferences. Those controls provide another answer-format measurement rather than a bias-free semantic probability.

For each semantic scenario $i$, the primary score is $\bar q_i=(q_{i,\mathrm{AB0}}+q_{i,\mathrm{AB1}})/2$. Label averaging is a transparent robustness summary; it is not a validated calibration method. UD and the individual assignments remain separate results. For input $F\in\{m,s,e\}$, the block contrast is
\begin{equation}
\Delta_{b,F}=\operatorname{mean}(\bar q_i\mid b,F=+1)-\operatorname{mean}(\bar q_i\mid b,F=-1),\qquad
\Delta_F=\frac{1}{24}\sum_b\Delta_{b,F}.
\end{equation}
Each marginalization averages over the other two inputs and section order. Effects are reported in percentage points. We also compute contrasts on the average candidate log-likelihood ratio to distinguish scale saturation from sign changes.

\subsection{Uncertainty and surrogates}
We use 10,000 percentile bootstrap resamples of the 24 numerical blocks. This holds the six authored commentary pairs and model set fixed, and quantifies only variation within this benchmark. Commentary-pair contrasts and leave-one-pair-out ranges provide separate wording checks. Four blocks per pair give limited replication; a six-pair bootstrap would not establish generalization to the universe of financial language. Profitability, annual-change, and gap-size subgroups are exploratory between-block comparisons and can also reflect assigned wording differences.

Ordinary least-squares surrogates predict $\bar q$ from momentum and sentiment, then add surprise, then section order and categorical pair/introduction indicators. Fitting leaves one earnings block out at a time; held-block scores never enter the training fit. We also leave one commentary pair out using the three financial inputs only. $R^2$ evaluates reproduction of model scores, not market accuracy. We report descriptive intervals and robustness checks without using a large collection of tests to declare a universal mechanism.

\section{Results}
\subsection{Cross-model input reliance}
\begin{table}[t]
\centering
\small
\begin{tabular}{llll}
\toprule
Checkpoint & Momentum & Tone & EPS surprise \\
\midrule
Qwen2.5-0.5B & 1.24 [0.816, 1.667] & 4.02 [3.089, 5.027] & $-0.15$ [$-0.506$, 0.192] \\
Qwen2.5-1.5B & 16.55 [15.224, 18.028] & 30.39 [26.558, 34.113] & $-0.19$ [$-0.444$, 0.065] \\
SmolLM2-1.7B & 6.27 [5.740, 6.822] & 6.46 [5.551, 7.355] & $-0.08$ [$-0.187$, 0.029] \\
TinyLlama-1.1B & 1.01 [0.859, 1.169] & 0.22 [$-0.134$, 0.587] & $-0.10$ [$-0.194$, $-0.003$] \\
Phi-3.5-mini & 20.77 [16.174, 26.089] & 45.01 [33.089, 57.099] & 6.57 [4.566, 8.772] \\
Qwen2.5-7B & 10.65 [6.702, 14.726] & 9.84 [5.359, 14.531] & 86.80 [81.027, 92.038] \\
\bottomrule
\end{tabular}
\caption{Primary positive-minus-negative input effects in percentage points (pp), averaging both A/B assignments on every case. Brackets show 95\% percentile intervals from 10,000 resamples of 24 earnings blocks. These are conditional benchmark intervals, not uncertainty about realized stock returns.}
\label{tab:main}
\end{table}

\begin{figure}[t]
\centering
\includegraphics[width=0.8\textwidth]{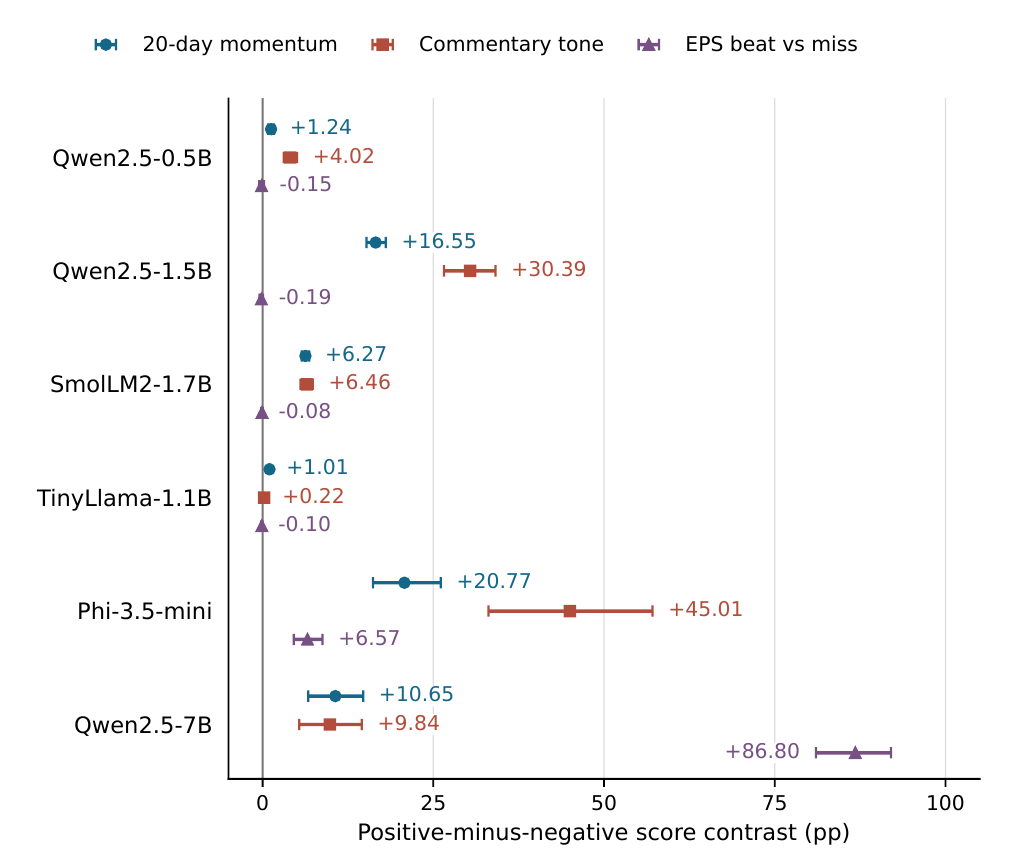}
\caption{Primary input contrasts. The three colored, shape-coded series share one horizontal scale; each row represents a checkpoint conversion. Intervals resample 24 earnings blocks conditional on the fixed conversion and wording set. Printed values give the signed contrasts in pp.}
\label{fig:main}
\end{figure}

Qwen2.5-0.5B: momentum 1.24 pp, tone 4.02 pp, and EPS surprise $-0.15$ pp. Its natural-word diagnostic accuracy is 55.2\%, and its A/B-averaged diagnostic accuracy is 50.0\%.

Qwen2.5-1.5B: momentum 16.55 pp, tone 30.39 pp, and EPS surprise $-0.19$ pp. Its natural-word diagnostic accuracy is 63.5\%, and its A/B-averaged diagnostic accuracy is 63.5\%.

SmolLM2-1.7B: momentum 6.27 pp, tone 6.46 pp, and EPS surprise $-0.08$ pp. Its natural-word diagnostic accuracy is 51.0\%, and its A/B-averaged diagnostic accuracy is 50.0\%.

TinyLlama-1.1B: momentum 1.01 pp, tone 0.22 pp, and EPS surprise $-0.10$ pp. Its natural-word diagnostic accuracy is 54.2\%, and its A/B-averaged diagnostic accuracy is 50.0\%.

Phi-3.5-mini: momentum 20.77 pp, tone 45.01 pp, and EPS surprise 6.57 pp. Its natural-word diagnostic accuracy is 84.4\%, and its A/B-averaged diagnostic accuracy is 77.1\%.

Qwen2.5-7B: momentum 10.65 pp, tone 9.84 pp, and EPS surprise 86.80 pp. Its natural-word diagnostic accuracy is 95.8\%, and its A/B-averaged diagnostic accuracy is 91.7\%.

Across this model set, magnitudes of the average signed tone contrast exceed magnitudes of the average signed EPS contrast in 5 of six checkpoints; momentum exceeds EPS by the same criterion in 5. This is a benchmark-specific ranking of directional responses. The strongest EPS contrast is observed for Qwen2.5-7B, at 86.80 pp; the smallest is Qwen2.5-1.5B, at $-0.19$ pp. Checkpoint differences combine architecture, pretraining, post-training, size, and conversion provenance. They do not identify a causal family or parameter-count effect.

The absolute value of a signed mean is not the mean absolute change. We therefore also report the absolute score movement for every matched input reversal (Table~\ref{tab:abs}), its EPS upper-tail magnitude, and the fraction of EPS pairs moving in the positive direction. A small signed mean can arise from cancellation and is not, by itself, evidence of complete insensitivity. Order-specific contrasts provide another check on averaging across prompt placement (Table~\ref{tab:order}). These are exploratory supplements to the unchanged primary signed estimand.

\subsection{Answer encoding and lexical variation}
\begin{table}[t]
\centering
\begin{tabular}{lrrrrr}
\toprule
Checkpoint & Label MAE & Flip (\%) & AB--UD & Order & Office \\
\midrule
Qwen2.5-0.5B & 12.06 & 43.0 & 11.43 & 3.26 & 7.50 \\
Qwen2.5-1.5B & 9.79 & 6.8 & 15.75 & 11.12 & 2.64 \\
SmolLM2-1.7B & 9.21 & 0.0 & 3.52 & 1.97 & 1.69 \\
TinyLlama-1.1B & 63.32 & 83.3 & 24.71 & 7.66 & 3.38 \\
Phi-3.5-mini & 17.65 & 17.7 & 9.46 & 6.01 & 6.02 \\
Qwen2.5-7B & 3.02 & 2.3 & 2.69 & 7.56 & 5.74 \\
\bottomrule
\end{tabular}
\caption{Forecast format and irrelevant-content controls. MAE columns are mean absolute score differences in pp. Label reversal and AB--UD use all 384 primary scenarios; order compares 192 matched cases; office compares 96 A/B-averaged cases. Flip is the share of A/B reversals changing whether the conditional positive-answer score exceeds 0.5.}
\label{tab:format}
\end{table}

Label reversal changes the question's answer mapping while preserving financial content. A/B averaging may reduce assignment preferences but cannot guarantee semantic invariance. The fully matched UD comparison tests another encoding; natural words provide a smaller additional check. A reversal of the sign or rank of a factor across these formats weakens any encoding-independent interpretation. Low requested-prefix mass further limits the meaning of scores conditional on those prefixes (Table~\ref{tab:mass}).

\begin{figure}[t]
\centering
\includegraphics[width=0.8\textwidth]{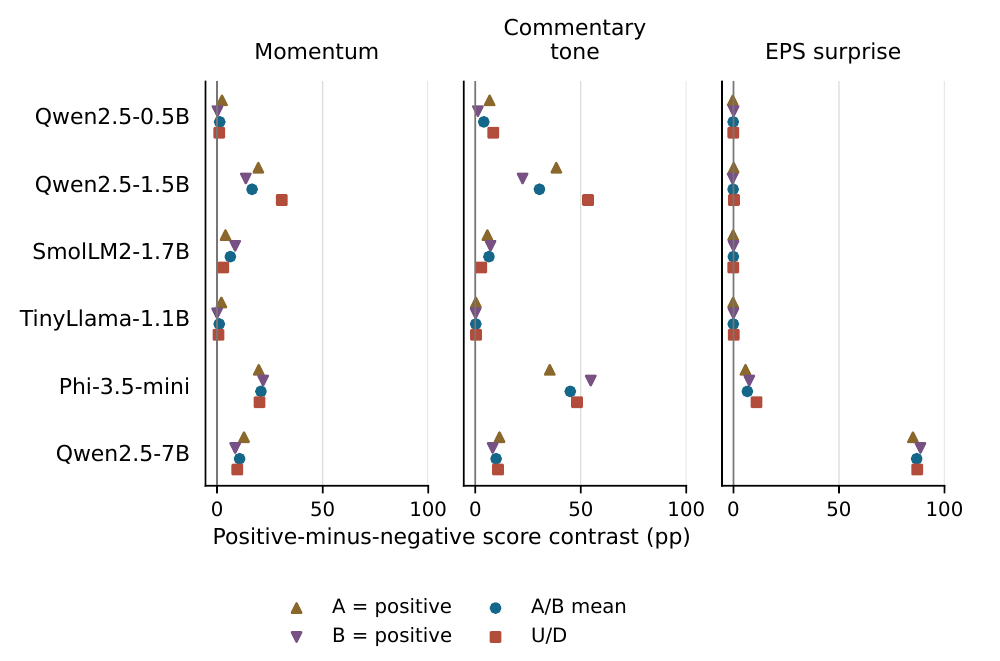}
\caption{Matched effects under AB0, AB1, their casewise score average, and UD on all 384 primary scenarios per model. Individual mappings are not treated as independent replications.}
\label{fig:enc}
\end{figure}

\begin{table}[t]
\centering
\begin{tabular}{lllll}
\toprule
Checkpoint & A=positive & B=positive & A/B mean & U/D \\
\midrule
Qwen2.5-0.5B & Tone & Tone & Tone & Tone \\
Qwen2.5-1.5B & Tone & Tone & Tone & Tone \\
SmolLM2-1.7B & Tone & Momentum & Tone & Momentum \\
TinyLlama-1.1B & Momentum & Tone & Momentum & Momentum \\
Phi-3.5-mini & Tone & Tone & Tone & Tone \\
Qwen2.5-7B & EPS & EPS & EPS & EPS \\
\bottomrule
\end{tabular}
\caption{Input with the largest absolute mean contrast in each encoding, on the same 384 scenarios. Ranking is descriptive and ignores interval overlap; negative effects can also have large absolute magnitude. Full signed effects are shown in Figure~\ref{fig:enc}.}
\label{tab:rank}
\end{table}

The wording results supply a direct check on the pilot's single-pair limitation. The six pairs broaden language coverage but share one authoring process. Pair-specific results and leave-one-pair-out ranges are therefore robustness evidence conditional on these templates, not population estimates of lexical invariance. The complete ranges are in the machine-readable summary. Natural-word comparisons use matched subsets rather than comparing unequal scenario samples (Table~\ref{tab:words}).

\subsection{Numerical recognition and commentary removal}
\begin{table}[t]
\centering
\begin{tabular}{lrrrrr}
\toprule
Checkpoint & A=beat & B=beat & Mean score & Mean accuracy & Words \\
\midrule
Qwen2.5-0.5B & 43.8 & 50.0 & 50.0 & 46.9 & 55.2 \\
Qwen2.5-1.5B & 60.4 & 70.8 & 63.5 & 65.6 & 63.5 \\
SmolLM2-1.7B & 50.0 & 50.0 & 50.0 & 50.0 & 51.0 \\
TinyLlama-1.1B & 50.0 & 50.0 & 50.0 & 50.0 & 54.2 \\
Phi-3.5-mini & 78.1 & 75.0 & 77.1 & 76.6 & 84.4 \\
Qwen2.5-7B & 90.6 & 94.8 & 91.7 & 92.7 & 95.8 \\
\bottomrule
\end{tabular}
\caption{Numerical EPS beat/miss accuracy (\%) on the same 96 scenarios per format. Mean score classifies the casewise average of A/B scores; mean accuracy averages the two assignment-specific accuracies. Words score the natural beat/miss prefixes. The methods estimate distinct quantities.}
\label{tab:diag}
\end{table}

The isolated EPS task has a known correct relation, whereas the next-day abnormal-return forecast does not have observed outcomes. Strong diagnostic performance can coexist with a small EPS forecast contrast: recognizing a fact and using it to form a forecast are different behaviors. Conversely, weak diagnostic performance prevents a small EPS contrast from being attributed solely to deliberate down-weighting of the surprise. Numerical accuracy remains answer-format dependent, so all assignments and natural words are retained.

Commentary removal is a context intervention, not a single-word sentiment flip. Its EPS effects are reported separately from the primary factorial average and should not be attributed exclusively to sentiment suppression. The office-color and section-order controls show whether nominally irrelevant changes move the score. The numerical subgroup results are given in Table~\ref{tab:subgroup}; the sample is too small and wording assignment too limited to establish universal differences between profits and losses.

\subsection{Simple surrogates and size comparisons}
A small incremental $R^2$ gain from EPS means that adding the surprise sign helps little in this particular surrogate. It does not prove that every computation inside the model is a two-cue rule, and omitted nonlinearities or interactions can matter. Pair and introduction indicators vary between blocks, so their $R^2$ increment can also absorb correlated numerical baseline variation; it is not an isolated causal template effect. The fully crossed order and label controls identify their specific prompt interventions more directly. The held-commentary-pair result is a harder wording transfer test than leaving out one numerical block. The three Qwen sizes permit a within-family descriptive comparison, but three checkpoints cannot support a fitted scaling law, and parameter count is confounded with training and post-training choices.

\begin{figure}[t]
\centering
\includegraphics[width=0.8\textwidth]{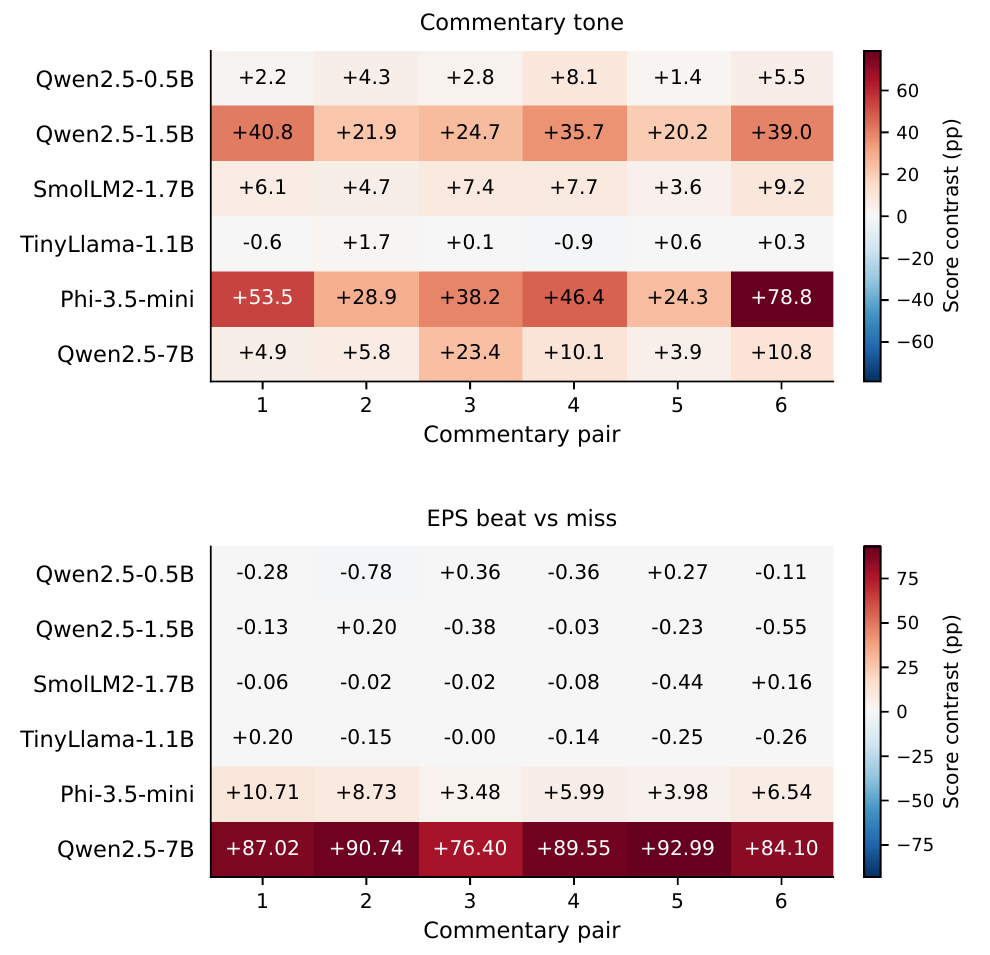}
\caption{Tone and EPS effects within each of six commentary pairs, using four numerical blocks per pair. The two stacked matrices have separate signed color scales and explicit pp values. Between-pair variation can include the numerical blocks assigned to each pair.}
\label{fig:pairs}
\end{figure}

\begin{figure}[t]
\centering
\includegraphics[width=0.8\textwidth]{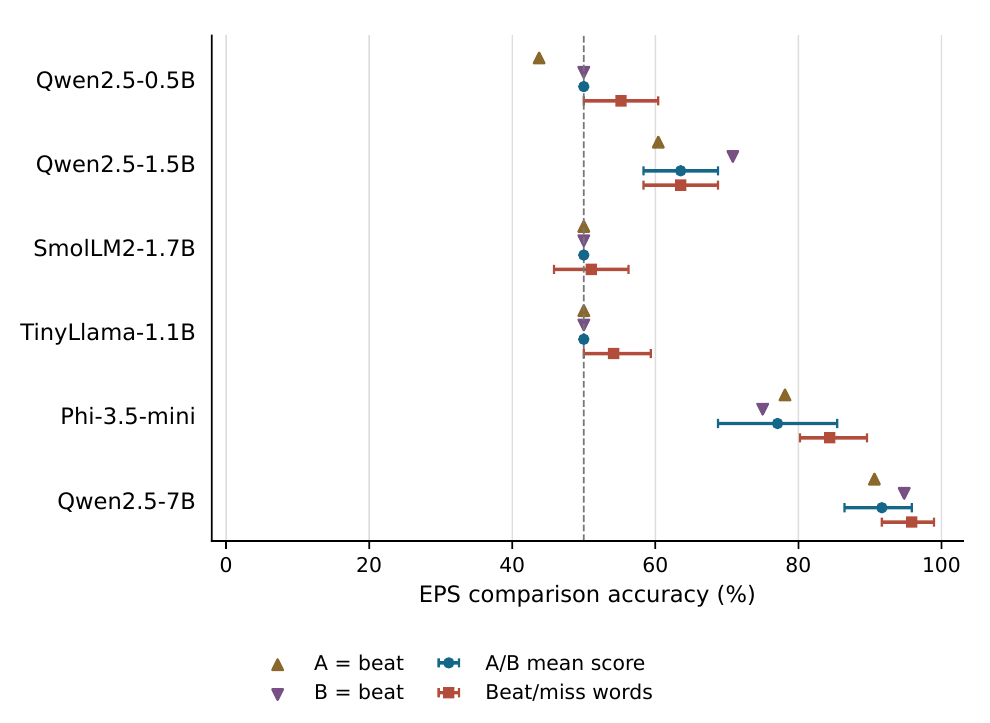}
\caption{Numerical beat/miss diagnostic on 96 balanced scenarios. The dashed line marks 50\% for this balanced task. Intervals on A/B-averaged and natural-word accuracy bootstrap 24 blocks; individual assignments are point estimates. The formats are distinct measurements; none is realized forecast accuracy.}
\label{fig:diag}
\end{figure}

\begin{table}[t]
\centering
\begin{tabular}{lrrrrr}
\toprule
Checkpoint & M+S & +E & +context & EPS gain & Held pair \\
\midrule
Qwen2.5-0.5B & 0.160 & 0.160 & 0.661 & $-0.000515$ & 0.107 \\
Qwen2.5-1.5B & 0.717 & 0.717 & 0.834 & $-0.000000$ & 0.702 \\
SmolLM2-1.7B & 0.767 & 0.767 & 0.772 & $-0.000005$ & 0.753 \\
TinyLlama-1.1B & $-0.045$ & $-0.045$ & 0.944 & 0.000033 & $-0.052$ \\
Phi-3.5-mini & 0.417 & 0.424 & 0.324 & 0.007589 & 0.414 \\
Qwen2.5-7B & 0.019 & 0.829 & 0.820 & 0.810390 & 0.830 \\
\bottomrule
\end{tabular}
\caption{Out-of-sample $R^2$ of linear surrogates of primary A/B-averaged scores. M+S uses momentum and tone; +E adds surprise; +context adds section order, commentary-pair and introduction indicators. First four numeric columns use leave-one-block-out fitting. Held pair uses M+S+E while leaving one of six commentary pairs out. Negative $R^2$ indicates worse performance than the pooled target-mean reference.}
\label{tab:r2}
\end{table}

\begin{figure}[t]
\centering
\includegraphics[width=0.8\textwidth]{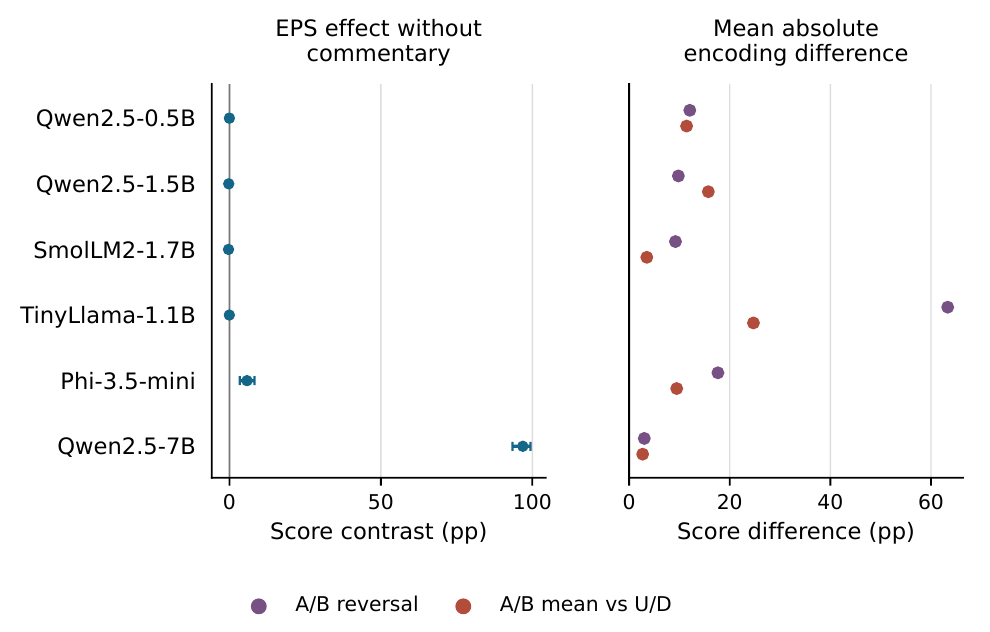}
\caption{Left: EPS effects after commentary removal, with block-bootstrap intervals. These cases use facts-first order only. Right: mean absolute score differences induced by answer mapping and by changing A/B to UD on matched primary cases.}
\label{fig:neutral}
\end{figure}

\begin{figure}[t]
\centering
\includegraphics[width=0.8\textwidth]{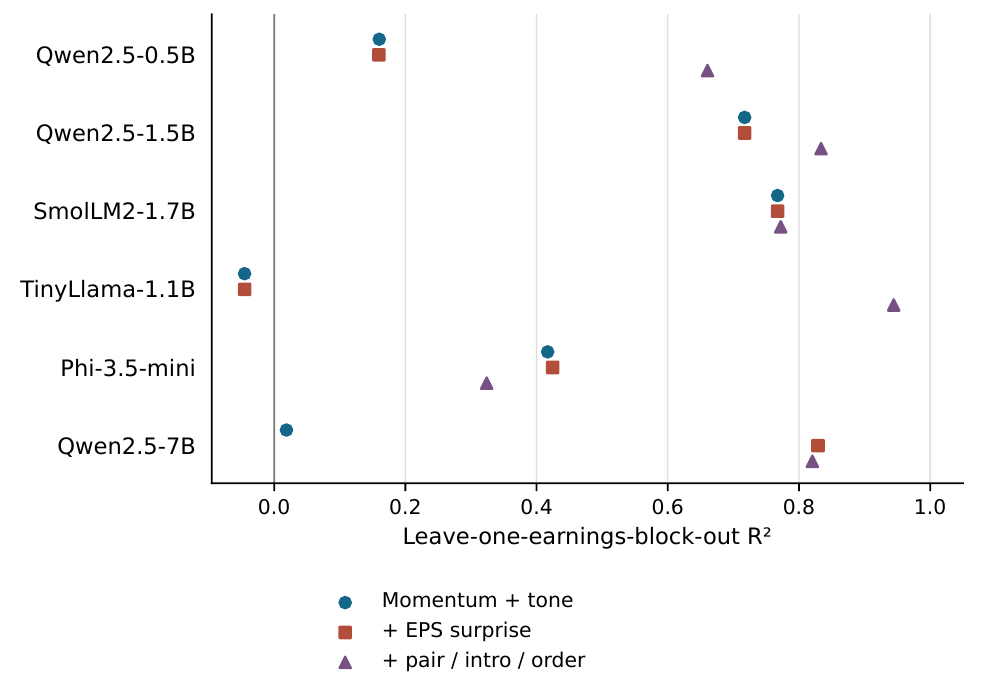}
\caption{Leave-one-block-out linear score reproduction. Adding EPS tests the incremental value of that manipulated input to this specific linear surrogate; context features can capture authored template differences.}
\label{fig:r2}
\end{figure}

\subsection{Bridge to the audited pilot}
Qwen2.5-0.5B has a mean absolute bridge difference of 15.64 pp, a maximum of 40.74 pp, and 22.9\% conditional decision flips.

Qwen2.5-1.5B has a mean absolute bridge difference of 6.21 pp, a maximum of 51.94 pp, and 8.3\% conditional decision flips. A targeted six-case investigation independently reproduces both the saved FP32 and downloaded-Q8 paths. Exporting the pinned HF source to native F16 gives a maximum score difference of 0.668 pp from the fresh FP32 path on these cases. Directly rebuilding Q8 from that F16 file differs from the downloaded Q8 conversion by as much as 46.136 pp. Only 73 of 291 stock tensor entries match byte-for-byte in the rebuilt file; tensor storage, source precision, preprocessing and converter choices can differ. Byte differences alone do not establish different training or an incorrect export. The measured outputs nevertheless show that these Q8 conversions are not interchangeable for the tested scores. This reinforces the scope of the main results: the specified conversion files, rather than a precision-invariant assessment of each named base checkpoint.

\begin{figure}[t]
\centering
\includegraphics[width=0.8\textwidth]{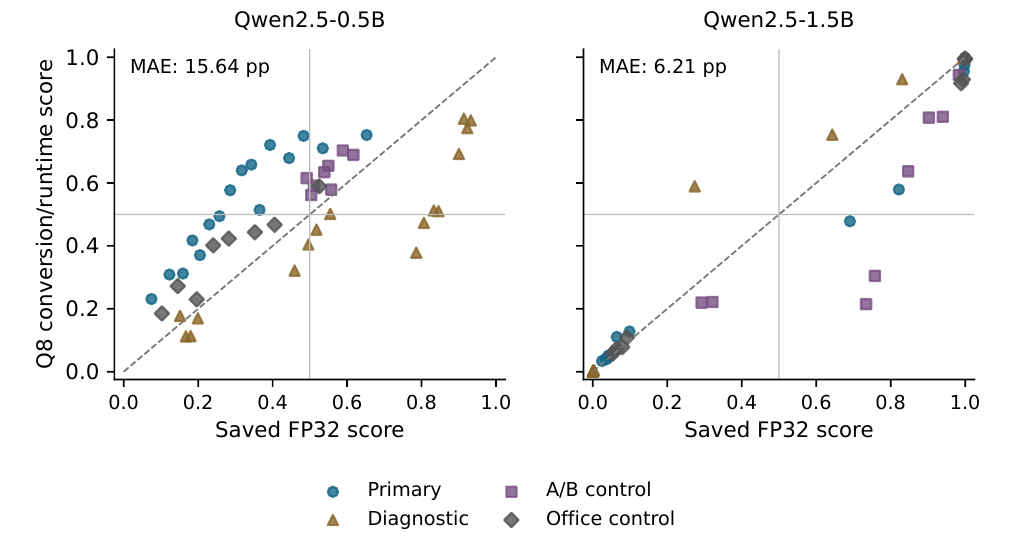}
\caption{Identical saved pilot prompts under the old FP32 setup and the new Q8 conversion/runtime. The dashed line marks equality and solid lines mark the conditional 0.5 threshold. Cases are selected across task types and saved score ranges, not randomly. This is a joint setup comparison, not a pure quantization experiment.}
\label{fig:bridge}
\end{figure}

The bridge uses 48 pilot cases per repeated Qwen checkpoint, selected across task/control variants and the saved score range. It compares conversion, precision, inference kernel, and execution together, rather than isolating quantization. Its maximum differences and flips are reported rather than assumed negligible. The material discrepancies prevent interpreting differences between the pilot and expanded estimates as purely a consequence of more diverse scenarios. These stress cases are not a random estimate of conversion error or diagnostic accuracy. The expanded analysis uses only the new common-runtime records and does not pool the pilot's full-precision outputs.

\section{Interpretation and limits}
On the primary A/B-averaged measure, the magnitude of the signed tone contrast exceeds that of the EPS contrast in 5 of six checkpoints; momentum exceeds EPS in 5. Using mean absolute matched changes instead, the corresponding counts are 5 and 5. The two summaries distinguish directional reliance from total score movement; both remain conditional on these supplied inputs and prompt formats.

Numerical recognition is heterogeneous. Qwen2.5-7B has the highest beat/miss word-prefix accuracy (95.8\%), whereas SmolLM2-1.7B has 51.0\%. The diagnostic has known answers; the forecast task has no observed return labels. High diagnostic accuracy therefore demonstrates performance on the numerical comparison task, not a profitable forecast or an identified internal reasoning mechanism.

Within the Qwen family, the 0.5B, 1.5B and 7B checkpoints have primary EPS contrasts of $-0.15$, $-0.19$, $+86.80$ pp and numerical word-prefix accuracies of 55.2\%, 63.5\%, 95.8\%. These are descriptive comparisons of three trained checkpoints, not an isolated causal effect of parameter count.

The EPS contrast remains positive under both individual A/B assignments and U/D for 2 of six checkpoints (Phi-3.5-mini, Qwen2.5-7B). A positive sign across encodings is a limited robustness check: magnitudes, decision thresholds and requested-prefix mass can still differ substantially. A/B averaging cannot guarantee semantic invariance.

A momentum-and-tone linear surrogate achieves held-block R-squared from $-0.045$ to 0.767; adding EPS changes it by $-0.000515$ to $+0.810390$. A two-cue account therefore fits some checkpoints better than others. These statistics reproduce model scores and do not estimate predictive value for actual market returns or exhaust nonlinear input dependence.

The controlled design identifies how specified inputs change answer-prefix scores for the tested conversion files and prompt formats. The expanded model and wording coverage makes cross-checkpoint and format heterogeneity measurable. Meaningful conclusions require attention to the financial contrasts, numerical diagnostics, and answer-format controls together. A result that appears under only one answer mapping cannot support a general claim about financial understanding. A small surprise contrast does not establish disregard for all fundamentals: actual EPS levels and annual changes remain fixed within the surprise intervention and vary between numerical blocks. Their broader influence is not separately identified by this primary estimand.

Several limits remain. First, these are six open-weight instruction/chat checkpoints at 0.49--7.61B parameters; no proprietary frontier model or dedicated long-deliberation model is tested. Second, all new inference uses Q8\_0 conversions, with some community provenance; accuracy and input reliance at other precisions may differ. Third, the disclosures are synthetic and short. Tone is deliberately stripped of new facts, whereas real narrative often contains relevant information. Fourth, changing consensus holds reported fundamentals fixed but changes another numerical fact, so the experiment estimates the effect of that expectation intervention under the prompt, not a pure conceptual operation inside the model. Fifth, the 24 blocks and six authored pairs limit uncertainty and subgroup interpretation. Sixth, prefix conditioning, prompt order, and label preferences may affect scores even after averaging assignments. The positive/beat alternative is described first in every format; letter reversal does not test semantic-option-order bias. Seventh, no stock returns, portfolios, transaction costs, or financial performance are measured. Eighth, the numerical gap is 2\% or 15\% of absolute actual EPS, rather than of consensus; cents rounding can alter its exact fraction. Finally, we use a direct-answer protocol without eliciting or intervening on chain-of-thought reasoning. The selected conversion checks show substantial output differences, so a full-precision replication is needed before treating these as precision-invariant properties of the named base models.

The evidence can support a statement about behavior on this benchmark, including patterns that vary by checkpoint and encoding. It cannot determine whether an LLM is ``really reasoning'' internally, or establish a market forecasting advantage. A next-stage study should freeze a prospective real-announcement sample, timestamp all information, add larger and different model families, and compare direct responses with deliberate reasoning and uncertainty-aware outputs under similarly complete encoding controls.

\section{Reproducibility and validation}
Public benchmark data and recorded scores are available in the Earnings Input Reliance Benchmark dataset\footnote{\url{https://huggingface.co/datasets/alinakhay/earnings-input-reliance-benchmark}}. The companion recorded-results explorer\footnote{\url{https://huggingface.co/spaces/alinakhay/earnings-input-reliance-explorer}} displays saved evaluations and performs no live inference. A compact GitHub showcase\footnote{\url{https://github.com/alinakhay/earnings-input-reliance-showcase}} provides an illustrative recorded-results demo and selected case comparisons.

The public dataset includes the benchmark prompts, numerical blocks, recorded predictions, matched primary comparisons, selected pilot-bridge records, and documentation of the scoring convention and conversion provenance. Model likelihoods are genuine local evaluations; no simulated model outputs or substituted heuristic predictions are used.

Independent validation reconstructs every score and prefix mass from saved log probabilities, verifies prompt/token hashes and candidate boundaries, checks exact semantic-cell coverage and numerical relations, and recomputes primary effects and diagnostic accuracies without the production analysis functions. Fresh uncached inference covers 12 cases per model spanning the primary experiment and controls. An initial runner read an unpopulated Python logits buffer; verification caught the empty scores, the reader was corrected to use native final-position logits, and all affected outputs were excluded and rerun. The exclusion receipt is preserved. The summary and figures are regenerated from completed valid records. These computational checks do not certify the absence of every possible methodological error.

\section{Conclusion}
Across six specified Q8 checkpoint conversions, the magnitude of the primary signed tone contrast exceeds that of EPS in 5 cases, while momentum exceeds EPS in 5. Those directional averages must be read alongside absolute matched changes, numerical diagnostic accuracy, and alternative encodings. The broader model and wording coverage reveals checkpoint-specific behavior rather than establishing a universal two-cue mechanism. The evidence measures input reliance under this protocol; it does not determine whether a model is internally reasoning or whether its forecast predicts market returns.

\bibliographystyle{plainnat}
\bibliography{references}

\appendix
\section{Complete authored commentary pairs}
\begin{table}[h]
\centering
\small
\begin{tabular}{cp{0.43\textwidth}p{0.43\textwidth}}
\toprule
Pair & Positive opinion & Negative opinion \\
\midrule
1 & This is an encouraging and impressive development; the outlook feels bright. & This is a disappointing and worrying development; the outlook feels bleak. \\
2 & I feel optimistic about this announcement. & I feel pessimistic about this announcement. \\
3 & My overall impression of this release is favorable. & My overall impression of this release is unfavorable. \\
4 & My reaction to this announcement is enthusiastic. & My reaction to this announcement is discouraged. \\
5 & I welcome this announcement and feel confident. & I regret this announcement and feel uneasy. \\
6 & This announcement leaves me upbeat about the shares. & This announcement leaves me downbeat about the shares. \\
\bottomrule
\end{tabular}
\caption{Six authored commentary pairs. The surrounding prompt explicitly states that the opinion contains no additional facts. These are wording variations from one construction process, not six independently sampled authors or validated human sentiment labels.}
\label{tab:pairs}
\end{table}

All three introductions and every numeric block are included in the public benchmark dataset linked in Section~6. Both section orders are scored for every primary factorial cell. Introduction and commentary assignment are balanced over blocks, not fully crossed within each block.

\section{Additional numerical and format results}
\begin{table}[h]
\centering
\begin{tabular}{lrrrr}
\toprule
Checkpoint & Momentum logit & Tone logit & EPS logit & Neutral EPS (pp) \\
\midrule
Qwen2.5-0.5B & 0.054 & 0.170 & $-0.006$ & $-0.02$ \\
Qwen2.5-1.5B & 0.783 & 1.413 & $-0.009$ & $-0.23$ \\
SmolLM2-1.7B & 0.326 & 0.331 & $-0.004$ & $-0.31$ \\
TinyLlama-1.1B & 0.047 & 0.042 & $-0.006$ & $-0.04$ \\
Phi-3.5-mini & 5.377 & 10.075 & 1.719 & 5.80 \\
Qwen2.5-7B & 3.468 & 3.524 & 15.635 & 96.86 \\
\bottomrule
\end{tabular}
\caption{Marginal effects on the average of the two answer-assignment log-likelihood ratios, and the EPS score contrast after removing commentary. Logit effects use a different scale from the main probability-score contrasts. Neutral EPS uses all 24 blocks at facts-first order only.}
\label{tab:logit}
\end{table}

\section{Exploratory precision and conversion investigation}
The converter is taken from the source distribution of the same pinned llama-cpp-python version. The source-distribution digest, new GGUF digests, case-level results, and tensor comparison are supplied. First-party and rebuilt files can differ in storage conventions; the rebuilt tied-output model omits the separate output tensor found in the stock file. No inference about training changes is made from file structure alone. The six cases are selected across observed bridge differences and do not estimate a random quantization error distribution.

\begin{table}[h]
\centering
\begin{tabular}{lrrrrr}
\toprule
Checkpoint & $|\Delta M|$ & $|\Delta S|$ & $|\Delta E|$ & EPS 95th & EPS $>0$ (\%) \\
\midrule
Qwen2.5-0.5B & 2.18 & 4.30 & 1.82 & 4.46 & 48.4 \\
Qwen2.5-1.5B & 16.55 & 30.39 & 1.35 & 3.26 & 50.0 \\
SmolLM2-1.7B & 6.27 & 6.47 & 0.86 & 2.07 & 42.7 \\
TinyLlama-1.1B & 1.09 & 1.00 & 0.60 & 1.38 & 44.3 \\
Phi-3.5-mini & 20.77 & 45.01 & 6.67 & 36.82 & 87.5 \\
Qwen2.5-7B & 10.69 & 10.94 & 86.80 & 99.99 & 100.0 \\
\bottomrule
\end{tabular}
\caption{Mean absolute matched score changes (pp), comparing 192 positive/negative input pairs per factor after A/B averaging. EPS 95th is the 95th percentile of absolute pair changes, not a confidence limit. EPS $>0$ is the share of beat-minus-miss pairs with a positive score change. These exploratory summaries detect cancellation hidden by a small signed mean.}
\label{tab:abs}
\end{table}

\begin{table}[h]
\centering
\begin{tabular}{llll}
\toprule
Checkpoint & Momentum & Tone & EPS surprise \\
\midrule
Qwen2.5-0.5B & 1.58 / 0.90 & 3.83 / 4.22 & $-0.29$ / $-0.01$ \\
Qwen2.5-1.5B & 13.61 / 19.50 & 33.58 / 27.20 & $-0.42$ / 0.05 \\
SmolLM2-1.7B & 6.82 / 5.71 & 6.64 / 6.29 & $-0.18$ / 0.03 \\
TinyLlama-1.1B & 0.90 / 1.12 & 0.27 / 0.17 & $-0.10$ / $-0.10$ \\
Phi-3.5-mini & 20.26 / 21.29 & 48.91 / 41.10 & 6.20 / 6.94 \\
Qwen2.5-7B & 8.94 / 12.37 & 10.77 / 8.91 & 86.88 / 86.72 \\
\bottomrule
\end{tabular}
\caption{Input contrasts (pp) in facts-first / facts-last section order, using the same 24 blocks in each order. These exploratory estimates preserve all other input matching and do not change the primary order-averaged estimand.}
\label{tab:order}
\end{table}

\begin{table}[h]
\centering
\begin{tabular}{lrrrrrr}
\toprule
Checkpoint & Loss & Profit & Decline & Rise & 2\% gap & 15\% gap \\
\midrule
Qwen2.5-0.5B & 0.21 & $-0.51$ & $-0.08$ & $-0.22$ & $-0.31$ & 0.01 \\
Qwen2.5-1.5B & $-0.44$ & 0.06 & $-0.48$ & 0.10 & $-0.28$ & $-0.10$ \\
SmolLM2-1.7B & $-0.15$ & $-0.00$ & $-0.01$ & $-0.15$ & $-0.16$ & 0.00 \\
TinyLlama-1.1B & $-0.15$ & $-0.05$ & $-0.15$ & $-0.05$ & $-0.02$ & $-0.18$ \\
Phi-3.5-mini & 2.50 & 10.65 & 6.08 & 7.06 & 4.79 & 8.35 \\
Qwen2.5-7B & 84.54 & 89.07 & 84.49 & 89.11 & 79.52 & 94.08 \\
\bottomrule
\end{tabular}
\caption{Primary EPS contrasts (pp) within profitability, annual EPS change, and surprise-size subgroups. Every subgroup has 12 earnings blocks. These exploratory comparisons span different numerical blocks and may also reflect differences in assigned wording; they do not isolate the causal effect of profitability or growth.}
\label{tab:subgroup}
\end{table}

\begin{table}[h]
\centering
\begin{tabular}{lrrrrrr}
\toprule
Checkpoint & Loss & Profit & Decline & Rise & 2\% gap & 15\% gap \\
\midrule
Qwen2.5-0.5B & 60.4 & 50.0 & 52.1 & 58.3 & 52.1 & 58.3 \\
Qwen2.5-1.5B & 52.1 & 75.0 & 62.5 & 64.6 & 64.6 & 62.5 \\
SmolLM2-1.7B & 50.0 & 52.1 & 45.8 & 56.2 & 52.1 & 50.0 \\
TinyLlama-1.1B & 56.2 & 52.1 & 52.1 & 56.2 & 54.2 & 54.2 \\
Phi-3.5-mini & 83.3 & 85.4 & 83.3 & 85.4 & 79.2 & 89.6 \\
Qwen2.5-7B & 91.7 & 100.0 & 95.8 & 95.8 & 91.7 & 100.0 \\
\bottomrule
\end{tabular}
\caption{Natural beat/miss prefix diagnostic accuracy (\%) within numerical subgroups; 48 scenarios and 12 blocks per subgroup. This known-answer task is distinct from return prediction. The numerical blocks differ between subgroups, so comparisons are descriptive.}
\label{tab:diagsub}
\end{table}

\begin{table}[h]
\centering
\begin{tabular}{llll}
\toprule
Checkpoint & Momentum & Tone & EPS surprise \\
\midrule
Qwen2.5-0.5B & 5.31 / 1.07 & 28.79 / 3.77 & 0.03 / $-0.12$ \\
Qwen2.5-1.5B & 33.40 / 16.82 & 50.91 / 30.85 & $-0.14$ / $-0.51$ \\
SmolLM2-1.7B & 19.99 / 6.23 & 18.83 / 6.52 & $-0.11$ / $-0.06$ \\
TinyLlama-1.1B & 0.12 / 0.95 & 2.31 / 0.18 & $-0.07$ / $-0.08$ \\
Phi-3.5-mini & 23.56 / 21.63 & 35.78 / 45.37 & 4.98 / 5.52 \\
Qwen2.5-7B & 10.37 / 12.95 & 9.26 / 13.38 & 85.01 / 83.38 \\
\bottomrule
\end{tabular}
\caption{Natural positive/negative word-prefix effects / matched A/B-averaged effects (pp). Each format covers exactly the same 192 scenarios: one loss and one profitable numerical block for each commentary pair, both section orders. Profitable cases were added in the documented control-coverage amendment. This test is descriptive and is not interchangeable with the 384-scenario primary estimand.}
\label{tab:words}
\end{table}

\begin{table}[h]
\centering
\begin{tabular}{lrrrrrr}
\toprule
Checkpoint & AB0 mass & top & AB1 mass & top & UD mass & top \\
\midrule
Qwen2.5-0.5B & 98.2 & 100.0 & 98.4 & 100.0 & 98.0 & 100.0 \\
Qwen2.5-1.5B & 99.8 & 100.0 & 99.7 & 100.0 & 99.9 & 100.0 \\
SmolLM2-1.7B & 94.6 & 100.0 & 93.7 & 100.0 & 94.0 & 100.0 \\
TinyLlama-1.1B & 38.5 & 88.3 & 57.9 & 100.0 & 11.9 & 35.2 \\
Phi-3.5-mini & 100.0 & 100.0 & 100.0 & 100.0 & 100.0 & 100.0 \\
Qwen2.5-7B & 100.0 & 100.0 & 100.0 & 100.0 & 100.0 & 100.0 \\
\bottomrule
\end{tabular}
\caption{Primary requested-prefix mass and first-token compliance (\%). Mass is the average unconditioned probability assigned to either requested prefix. Top is the share whose highest-logit next token starts a requested candidate. Both are distinct from semantic comprehension or calibrated market probability.}
\label{tab:mass}
\end{table}

\begin{table}[h]
\centering
\small
\begin{tabular}{lrrrr}
\toprule
Pilot case & FP32 & Rebuilt F16 & Rebuilt Q8 & Stock Q8 \\
\midrule
b02\_m1\_s-1\_e1\_t0\_o0\_labels0 & 55.76 & 55.09 & 66.96 & 57.88 \\
b08\_e1\_diagnostic1\_reversed & 15.13 & 15.27 & 20.32 & 17.75 \\
b00\_m1\_s-1\_e-1\_t0\_o1 & 39.31 & 39.83 & 44.58 & 72.10 \\
b10\_e1\_diagnostic1 & 80.62 & 80.59 & 85.73 & 47.32 \\
b07\_e1\_diagnostic1 & 84.54 & 84.84 & 86.12 & 51.04 \\
b00\_e-1\_diagnostic1 & 78.57 & 78.67 & 83.97 & 37.83 \\
\bottomrule
\end{tabular}
\caption{Conditional positive/beat candidate scores (\%) for six selected Qwen0.5 pilot stress cases. The rebuilt F16 and Q8 files use the same pinned HF source; stock Q8 is the downloaded first-party conversion used in the main experiment. Cases span tasks and are selected after observing bridge differences; this is an exploratory precision/provenance investigation, not an error-rate sample or a new full-precision factorial experiment.}
\label{tab:bridge}
\end{table}

\section{Exact prompt and candidate convention}
The shared system instruction is: \emph{You are a financial analyst. Follow the requested answer format.} Official model-specific chat templates wrap it and the user message. A primary user message includes the block introduction, all three disclosure sections, and:

\begin{quote}
\emph{Forecast the sign of the stock's abnormal price reaction during the next trading day following this release. Answer exactly A for a positive reaction or B for a negative reaction. Output the requested answer only.}
\end{quote}

AB1 swaps the A/B mapping; UD uses U/D. Words replace the candidate labels with positive/negative. The EPS task asks:

\begin{quote}
\emph{Did actual EPS beat or miss consensus? Answer exactly A for beat or B for miss. Output the requested answer only.}
\end{quote}

Diagnostic AB1 reverses A/B; words use beat/miss. Candidate strings are appended directly to the template's generation prefix, with no invented leading space. Their suffix token IDs are derived from the joint prompt-plus-candidate encoding. This matters especially for SentencePiece tokenizers. Likelihoods include each suffix token and exclude an end-of-answer token.

\section{Artifacts and protocol chronology}
The pilot was designed and run before this expansion and had already revealed answer-encoding sensitivity. The expanded primary protocol therefore incorporates controls suggested by that observed limitation. Its original 1,824 cases are internally timestamped before expanded results, not a pilot-independent public preregistration. The later 192-case control-coverage addition follows a documented audit of the first-block selection rule; it is not presented as preregistered confirmation. The public benchmark dataset retains the original frozen prompts, expanded scores, protocol, and coverage amendment. The new benchmark is analyzed as a separate dataset under one new runtime. No pilot hypothesis is relabeled as an unseen confirmation.

\end{document}